# Politicians vs ChatGPT.
# A study of presuppositions in French and Italian political communication

*Davide Garassino* (Zürcher Hochschule für Angewandte Wissenschaften), *Viviana Masia* (Università Roma Tre), *Nicola Brocca* (Universität Innsbruck) & *Alice Delorme Benites* (Zürcher Hochschule für Angewandte Wissenschaften)

davide.garassino(at)zhaw.ch, viviana.masia(at)uniroma3.it, nicola.brocca(at)uibk.ac.at, alice.delormebenites(at)zhaw.ch

## Abstract

This paper aims to provide a comparison between texts produced by French and Italian politicians on polarizing issues, such as immigration and the European Union, and their chatbot counterparts created with ChatGPT 3.5. In this study, we focus on implicit communication, in particular on presuppositions and their functions in discourse, which have been considered in the literature as a potential linguistic feature of manipulation. This study also aims to contribute to the emerging literature on the pragmatic competences of Large Language Models. Our results show that, on average, ChatGPT-generated texts contain more questionable presuppositions than the politicians' texts. Furthermore, most presuppositions in the former texts show a different distribution and different discourse functions compared to the latter. This may be due to several factors inherent in the ChatGPT architecture, such as a tendency to be verbose and repetitive in longer texts, as exemplified by the occurrence of political slogans mainly formed by change-of-state verbs as presupposition triggers (e.g., *dobbiamo costruire il nostro futuro*, 'we must build our future').

## Keywords

Implicit communication, Presupposition, AI-generated language, Political communication, French, Italian

"I expect ai to be capable of superhuman persuasion well before it is superhuman at general intelligence, which may lead to some very strange outcome".
(Sam Altmann, X, 25.10.2023)

## 1 Introduction

The aim of this paper is to provide an exploratory investigation of the use and discourse functions of presuppositions in speeches produced by politicians and texts generated by ChatGPT (OpenAI 2023). Presuppositions are usually considered potentially persuasive implicit content (Sbisà 2007; Lombardi Vallauri 2016 and 2021), whose investigation can help to explore and understand the use of manipulative strategies in political communication. The choice of comparing politicians' speeches and Artificial Intelligence (henceforth, AI)-generated texts is due to the recent worldwide success of Large Language Models (LLMs) in general, and ChatGPT, in particular. In light of the proliferation of generated texts, it seems to us the right time to assess their textual and pragmatic properties as well as to estimate their potential for spreading fake news and questionable content, which is one of the most feared aspects of LLMs in public discourse and in the expert community (see Goldstein, Chao, Grossman et al. 2023 for a recent discussion).

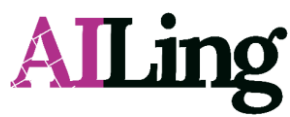

The main contribution of our analysis is to show that, on average, ChatGPT-generated texts contain more questionable presuppositions than the original politicians' speeches. Furthermore, most presuppositions in ChatGPT texts have a different distribution and different discourse functions compared to the politicians' speeches. This may be due to several factors, including the fact that longer texts generated by this LLM tend to be repetitive and biased, as they select and repeat (more or less) empty slogans, mainly using change-of-state verbs as presupposition triggers (such as *dobbiamo costruire il nostro futuro*, 'we must build our future').

The paper is structured as follows: in Section 2, we build on the key concepts of persuasion and manipulation to zoom in on the persuasive and/or manipulative power of AI tools. Section 3 presents our research questions. The working corpus, which was obtained by collecting authentic speeches of French and Italian politicians as well as texts generated by ChatGPT 3.5., is described in Section 4. A quantitative corpus-based analysis is provided in Section 5. Section 6 discusses the main features of the generated texts and the general implications of the analysis for political communication, while Section 7 addresses some limitations of the present study.

## 2 Background
### 2.1 Persuasion, manipulation and artificial intelligence

As evidenced by the message quoted above from Sam Altmann (the CEO of OpenAI), persuasion is considered a risk associated with the development of AI systems. Likewise, numerous other sources, including both research and press, have recently used the terms *persuasion* and *manipulation*, often interchangeably.

Elsewhere, however, the two terms have been kept separate; for example, some work in the field of language and communication evolution has proposed a distinction based on mutual costs and benefits for both speakers and hearers (Reboul 2011 and 2017).

If we acknowledge that human communication is naturally oriented towards argumentation and persuasion, i.e., towards changing the mental states and beliefs of others, as suggested by Mercier and Sperber (2017), among others, persuasion can be considered as a ubiquitous feature of human communication. More precisely, following Reboul (2017), persuasion can even be considered *cooperative* when it is beneficial for the speaker and neutral or beneficial for the hearer. However, persuasion can be *exploitative* when it is only beneficial for the speaker but costly for the hearer. In this paper, we use the term *persuasion* when referring to cooperative persuasion and *manipulation* when referring to exploitative persuasion. With the recent success of LLMs, there has been growing concern about the manipulative uses of AI tools (see, for example, Burtell and Woodside 2023 for a recent review). Manipulative uses can involve, for example, biases related to gender and sociopolitical framing (Bender, Gebru, McMillan-Major and Shmitchell 2021), but also the spread of fake news and disinformation (Buchanan, Lohn, Musser and Sedova 2021).

As suggested by Bai, Voelkel, Eichstaedt and Willer (2023), texts generated by current LLMs are able to persuade people, even when polarized issues are at stake (such as a ban on assault weapons or a carbon tax, among others).

In particular, LLMs such as ChatGPT 3.0, 3.5 and 4.0 have been recognized as being able to "maximise the benefits of propagandists in terms of text quality and reach" (Goldstein, Sastry, Musser et al. 2023). Goldstein, Chao, Grossman et al. (2023) report the results of a perception experiment in which participants indicated their agreement (or disagreement) with a controversial or blatantly false thesis contained in articles generated either by human propagandists or by ChatGPT 3.0. Interestingly, on average, participants showed similar levels of high agreement with both types of texts. When the ChatGPT texts were enhanced with human intervention (e.g., editing the prompt and curating the output), they proved comparable to original propaganda and in some cases even outperformed the texts produced by human propagandists. Recent research has also shown that, despite efforts by OpenAI and Google to prevent their chatbots from producing false and offensive content, it is possible (and easy) to bypass such content filters and produce offensive content (Zou, Wang, Carlini et al. 2023).

According to our distinction between persuasion and manipulation, some uses of AI tools, particularly chatbots, are (or can be) potentially manipulative, that is, they can be beneficial to propagandists and costly to readers. Other lines of research have also shown that humans (and machines alike, Chaka Chaka 2023) are usually unable to distinguish between texts generated by human agents and texts generated by ChatGPT (Kreps, McCain and Brundage 2022).

In light of this, countering manipulative content seems difficult, and even more so when we consider that people have limited time and cognitive resources to devote to processing tasks, making their adherence to a strategy of *rational ignorance* (Downs 1957) a viable choice in most contexts. In other words, people tend to simply ignore gathering additional information or checking the sources of dubious information because it is a costly operation with no immediate, tangible benefit (see also Goldstein, Sastry, Musser et al. 2023).

The remainder of this paper presents a corpus-based study of potentially manipulative political communication by focusing on a linguistic feature, presupposition, which we will explore in Section 2.2.

## 2.2 Presupposition as a linguistic trait of manipulation

Implicit communication appears to have a strong link with manipulation in language, since it hides speakers' manipulative intentions, thus making them less obvious and easy to detect. This is particularly true in domains such as advertising and political discourse (Brocca, Garassino and Masia 2016; Lombardi Vallauri 2019). As already hinted at in Section 1, in this paper we will mostly zoom in on a common strategy of implicit meaning, namely presupposition.

Over the last decades, there has been a long debate on the manipulative power of presuppositions and on how they might allow conveying some contents below the level of critical attention (Sbisà 2007; Lombardi Vallauri 2009). From a Stalnakerian perspective, a presupposition can be characterized as taken for granted information (Stalnaker 1973), that is, knowledge the speaker assumes to be already in the *common ground* of the conversation (Stalnaker 2002). In previous works, such as Ducrot (1972; 1984), presupposition has been defined as a linguistic act having the function of restraining the speaker's possibilities of orienting the subsequent discourse. In line with the mainstream literature, we take common ground to refer to the presumed background information shared by participants in a conversation (Stalnaker 2002: 701).

Differently from other types of implicit meaning (i.e. conversational implicatures, among others), presuppositions are typically *activated* by means of dedicated linguistic categories or structures, known as *presupposition triggers*. Very common presupposition triggers are definite descriptions (e.g., the sharp knife), change-of-state verbs (e.g., stop, continue, etc.), adverbial subordinate clauses (e.g., when Jane went to England…), iterative and other focusing adverbs (e.g., also, too, again, etc.), factive predicates (e.g., I *blame* John for having left me without an explanation; *it's wonderful* that you could be given this opportunity!), among others (more exhaustive lists of triggers can be found in Kiparsky and Kiparsky 1971; Sbisà 2007; Lombardi Vallauri 2009).

A distinctive trait of presupposition behavior in discourse is its resistance to negation (or other logical operators), in that its truth value remains constant when the wider sentence in which it occurs is negated. So, for example, in *It is not true that John stopped smoking*, to be negated here is not that John used to smoke but only that he has presently interrupted this activity; the sentence thus maintains its presupposition untouched.

The earlier experimental literature on presupposition processing (Loftus 1975; Langford and Holmes 1979; Schwarz 2015, among others) has evidenced a strong likelihood of presuppositions being decoded in a *shallow way* (Masia, Garassino, Brocca and de Saussure 2023), namely without thoroughly attending to all its details. In so doing, presuppositions prove to be particularly effective in making some content escape receivers' epistemic vigilance (Sperber, Clément, Heintz et al. 2010; cf. also Lombardi Vallauri 2021).

In certain contexts of language use, presuppositions often happen to correlate with contents which are not entirely objective or universally true (Garassino, Brocca and Masia 2022). In fact, although their most typical function is to convey contents assumed to be known in advance by all participants in a communicative exchange, their discursive and processing behavior may also be exploited to convey unshared and non-objectively verifiable contents, that is, contents that are more often than not the manifestation of the speaker's subjective opinion or view on something. At times, this may result in the use of presuppositions which are strongly tendentious or blatantly false. Following Lombardi Vallauri (2019), Garassino, Brocca and Masia (2022) and Cominetti, Gregori, Lombardi Vallauri and Panunzi (2022), we can refer to these content types as *non-bona fide* true or as *potentially manipulative presupposition* (henceforth, PMP). An example of PMP is given below:[1]

---

[1] Examples (1) and (2) were extracted from the IMPAQTS corpus, collected within the PRIN project IMPAQTS: *Implicit Manipulation in Politics - Quantitatively Assessing the Tendentiousness of*

(1) *Questa riflessione sulla mancanza di libertà in Italia* (Alfonso Bonafede - Movimento Cinque Stelle)[2]
    'This reflection on the lack of the citizens' freedom in Italy' (our transl.)

Conversely, a *bona fide* true presupposition is exemplified in (2).

(2) *Per l'assenza* [...] *del turismo internazionale* (Dario Franceschini, Democratic Party)[3]
    'Because of the absence of international tourism' (our transl.)

Now, while the lack of tourism due to the lockdowns in 2020 can be taken to be an overall tangible fact (potentially backed by objective data), that Italian citizens are not free is more a political slogan than a fact. However, it must be highlighted that the context in which a presupposition is used necessarily impinges on an evaluation of its nature as *bona fide* true or non-true.

In texts with a possible manipulative aim (such as the already mentioned advertising and political propaganda), the use of presupposition strategies seems to be particularly oriented towards the encoding of specific types of contents (henceforth, we will refer to these latter as the *discourse functions* of presupposition). Recent research conducted on corpora of political language (Garassino, Masia and Brocca 2019 and Garassino, Brocca and Masia 2022), including both traditional speeches and messages posted by politicians on Twitter or other Social Networks, showed that the content types being more commonly associated to presupposition strategies are criticism, self-praise, praise-of-others and stance-taking. Following the working definitions proposed in Garassino, Masia and Brocca (2019), we will couch these discourse functions as follows:

- CRITICISM: attack directed to other politicians or political groups with respect to some specific issue.
- SELF-PRAISE: praise centered on the speaker herself or her political group for some achievement or behavior.
- PRAISE-OF-OTHERS: praise or appreciation shown to other people for their achievements or for supporting a certain political party.
- STANCE-TAKING: opinion or attitude of a politician towards a specific issue.

Examples of such discursive functions of presupposition strategies are given in (3).

---

(3)
CRITICISM
a. Noi demoliremo *l'euromostro* (Meloni, T1)[4]
'We'll demolish the Euromonster'
Presupposition: There is a Euromonster.

SELF-PRAISE
b. Nous *poursuivrons* sans relâche notre travail pour l'inclusion de nos compatriotes en situation de handicap (Macron, T2)
'We will relentlessly continue our work for the inclusion of our compatriots with disabilities'
Presupposition: We have been working so far to include our fellow citizens with disabilities.

PRAISE-OF-OTHERS
c. Heureuse d'être ici, chez nous, sur *cette terre si française, populaire et patriote*, du bassin minier (Le Pen, T2)
'I'm happy to be here, at home, in this land that is so French, so popular and so patriotic, in the Mining Basin.'
Presupposition: There is a land that is so French, so popular and so patriotic.

STANCE-TAKING
d. Per *riportare questo paese sui giusti binari* (Zingaretti, T1)
'To get this country back on track'
Presupposition: This country is not currently on track.

The definite noun phrase *l'euromostro* in (3a) presupposes the existence of a Euromonster with which Giorgia Meloni expresses her disappointment and discontent towards European monetary policy. The change-of-state verb *poursuivre* in the French example in (3b) presupposes that the French President Macron and his government have already been active in helping people with disabilities. The definite noun phrase *cette terre si française, populaire et patriote* in (3c) conveys as already shared information that the Mining Basin in Northern France is a truly French, popular and patriotic region. Finally, in (3d), the change-of-state verb *riportare* in the Italian example presupposes the politician's intention to bring Italy back on track, thus expressing his stand on the actual condition of the country.

The research so far undertaken on the manipulative power of presuppositions (see Lombardi Vallauri and Masia 2014, among others) has not yet thoroughly delved into impact differences between trigger types. Notably, it cannot be said with certainty whether a trigger presupposes more strongly or is more manipulative than another. However, some recent experimental findings on the processing of presupposed contents have pointed to different neural mechanisms underlying the mental encoding of presuppositions triggered by subordinate clauses as compared to definite descriptions (Masia, Canal, Ricci et al. 2017) and those related to definite descriptions as compared to change-of-state verbs (Domaneschi, Canal, Masia et al. 2018; Masia, Garassino, Brocca and de Saussure 2023). Although interesting, these data do not allow drawing conclusive generalizations on whether increasing processing costs should be taken as indexing a stronger manipulative impact than triggers imposing less taxing mental operations.

---

[4] The labels T1 and T2 refer to the texts collected for the corpus used in this study. Since our corpus contains two speeches for each politician and two ChaGPT-generated texts for their chatbot versions, in this paper we use the following reference scheme: (Politician name, T1 or T2) and (Politician name_ChatGPT, T1 or T2).

## 3 Research questions

The main research question that we intend to address in this paper is: *Do the chatbot versions of French and Italian politicians use presuppositions in a similar way to their real-life counterparts?* This question can be further split into three sub-questions:

Q1. Frequency: Does ChatGPT produce, on average, texts with more or less PMPs in comparison with real texts produced by politicians?
Q2. Form: Is the form of the PMPs (i.e. the triggers that activate them) in ChatGPT's texts similar to that in the politicians' texts?
Q3. Function: Is the discourse function of PMPs in ChatGPT's texts similar to that of the politicians' texts?

## 4 The corpus: Politicians' and ChatGPT-generated texts

In order to answer Q1-Q3, we rely on an exploratory corpus-based study. The corpus used for the analysis consists of texts generated by politicians and ChatGPT. The texts produced by politicians are speeches given by French and Italian politicians between 2018 and 2022 during rallies in national election campaigns. The French politicians selected were Emmanuel Macron, at that time President of the French Republic and Secretary of the *La République en Marche!* party, and Marine Le Pen, Secretary of the *Rassemblement National* party. The Italian politicians selected were Nicola Zingaretti, Secretary of the *Partito Democratico*, and Giorgia Meloni, Secretary of the *Fratelli d'Italia* party. All politicians led major parties in their respective countries and represented opposing political orientations. Their speeches tackle similar topics, primarily focusing on the European Union and immigration (see the online attachments in OSF for the Corpus).[5]

The chatbot texts were generated using ChatGPT (OpenAI 2023), version 3.5, which was freely available at the moment of the collection, on July 1, 2023. To create an effective prompt, Ekin's (2023) guidelines were followed as well as the *persona prompt pattern* strategy (see 3a), whose motivation is to give "the LLM a "persona" that helps it select what types of output to generate and what details to focus on" (White, Fu, Sandborn et al. 2023: 5).

---

[5] The attachments are available at https://osf.io/yk7be/ (DOI 10.17605/OSF.IO/YK7BE).

The prompts we created for generating the texts rely on the same structure (see an example in (4)). Firstly, the LLM was requested to adopt the politician's identity and a realistic setting (place, time) was also provided, (4a). Secondly, a brief excerpt from an actual speech followed (4b). Thirdly, in (4c), the length of the text was added and two internal parameters of the LLM were also manually adjusted (but see the discussion in Section 7): the *temperature* of the model was increased to enhance the creativity of the output, and its *diversity penalty* was decreased to reduce repetition. Finally, in (4d), the prompt concludes with some additional style guidelines and suggestions on the selections of the discourse topics. The prompt was written in English, but the LLM was asked to produce the output in French or in Italian (see 4c).

(4) a) Imagine that you are Nicola Zingaretti, Secretary of the Italian *Partito Democratico*, delivering a speech at a rally in Milan during the 2019 European elections

   b) Continue the following speech: "[Insertion of a few lines of a Nicola Zingaretti's speech]".

   c) Write approximately 5,500 characters in Italian.
   Use a *temperature* of 0.8.
   Use a *diversity_penalty* of 1.5.

   d) Be persuasive and specific in your speech, emphasizing the importance of a cultural and democratic battle for the well-being of Italy as well as the importance of the European Union for the Italian economy and society.

To assess the clarity of the prompt, we consulted ChatGPT. ChatGPT informed us that the prompt was detailed enough and the required length of the output appropriate, but, due to internal limitations of the model, the generated responses are shorter than 5,500 characters (see Table 1). The prompt was entered directly into the chat interface. A test was run with the same prompt using the API in R (v. 4.2.2; R Core Team 2022), which gave comparable results (see OSF for the code).

Table 1: Length of real politicians' speeches and texts generated by ChatGPT 3.5.

|  | Politicians | | ChatGPT | |
| --- | --- | --- | --- | --- |
|  | T1 | T2 | T1 | T2 |
|  | words / tokens | words / tokens | words / tokens | words / tokens |
| M. Le Pen | 1,024 / 5,508 | 1,015 / 5,153 | 614 / 3,262 | 672 / 3,577 |
| E. Macron | 1,087 / 5,731 | 971 / 5,330 | 679 / 3,753 | 706 / 3,809 |
| G. Meloni | 1,029 / 5,267 | 1,048 / 5,758 | 765 / 4,011 | 624 / 3,409 |
| N. Zingaretti | 1,012 / 5,328 | 992 / 5,368 | 560 / 3,071 | 559 / 3,120 |

## 5 Results
## 5.1 Annotation and interrater agreement[6]

The texts were annotated by the first and the second author in relation to the independent variables *Presupposition Trigger* and *Discourse Function* (see Section 2).

The first step in the annotation process was to identify which presuppositions in the texts counted as PMPs. To accomplish this, the annotators first identified these items independently and then reviewed them collectively during a consensus through negotiation meeting (Lowen and Plonsky 2016). The annotators compared each case in their own list of items identified as PMPs: if the occurrence was already present in both lists, it was automatically added to the final selection, otherwise it was discussed until perfect agreement was found. If one of the two annotators still had doubts, the item was discarded. Note also that no formal analysis based on interrater agreement indices (see below) was carried out at this stage, since PMP occurrences were only identified in texts but not annotated in relation to some other categories (e.g., non-PMP occurrences).

After agreeing on the PMP items, the annotators independently annotated the *Presupposition Trigger* and *Discourse Function* variables. As the annotation of pragmatic categories such as our *Discourse Function* can be notoriously difficult (see the discussions in Spooren and Degand 2010; Grisot 2017; Garassino, Brocca and Masia 2022), the annotators relied on a codebook (see OSF). The codebook aimed to reduce ambiguity, for example between categories such as Criticism and Stance-taking (Section 2), by limiting Criticism to explicit criticism or attacks addressed to political rivals, while including in Stance-taking implicit criticism directed at unspecified targets.

---

Agreement between the annotators was calculated using two indices: Cohen's k and Gwet's AC1.[7] The former is one of the most widely used measures of interrater agreement (see Loewen and Plonsky 2016: 28), while the latter has recently gained popularity as a potential alternative to Cohen's *k* (see Hoek and Scholman 2017).[8] As noted by Loewen and Plonsky (2016: 28), "there is no agreed standard for a kappa score to be considered acceptable". For example, according to Landis and Koch (1977), values between 60% and 80% could be considered to indicate substantial agreement, while more conservative views suggest that values of around 67% may only indicate a tentative conclusion (Artstein and Poesio 2008: 576). In any case, values above 80% are generally considered to show a high level of agreement (Loewen and Plonsky 2016: 90). Similar to Cohen's *k*, there are no agreed standards for Gwet's AC1. According to Hoek and Scholman (2017: 7), when assessing the results, "it might also be warranted that the guidelines for what constitutes satisfactory agreement are slightly stricter for AC1" compared to Cohen's *k*.

Table 2: Interrater agreement indexes across politicians and their chatbot versions (means and standard deviations).

|  | Politicians | | ChatGPT | |
|---|---|---|---|---|
|  | Cohen's *k* | Gwet's AC1 | Cohen's *k* | Gwet's AC1 |
| Presupposition Trigger | μ = 89.5 sd= 4.65 | μ = 95.5 sd= 2.38 | μ = 95.75 sd= 3.77 | μ = 97 sd= 2.58 |
| Discourse Function | μ = 65.5 sd = 13.91 | μ = 79 sd= 9.12 | μ = 65.25 sd = 5.85 | μ = 83.5 sd= 4.65 |

---

[7] The R package *irr* (v. 0.84.1; Gamer and Lemon 2019) was used to calculate the two indices.

[8] Some researchers suggest that the Gwet's AC1 index may be more suited for linguistic categories and their distribution in corpus data than Cohen's *k*, as the former is more robust to "variability in the distribution of categories" (Hoek and Scholman 2017: 7; also see Hoek, Scholman and Sanders 2021). However, the fact that Gwet's AC1 can truly be an alternative to Cohen's *k* remains debated (Vach and Gerke 2023). For this reason, we decided to include both measures in this paper.

Table 3: Interrater agreement indexes (values for each politician and each chatbot version).

| | Politicians | | | | ChatGPT | | | |
| | Presupposition Trigger | | Discourse Function | | Presupposition Trigger | | Discourse Function | |
| | Cohen's $k$ | Gwet's AC1 | Cohen's $k$ | Gwet's AC1 | Cohen's $k$ | Gwet's AC1 | Cohen's $k$ | Gwet's AC1 |
|---|---|---|---|---|---|---|---|---|
| M. Le Pen | .94 | .98 | .63 | .75 | .97 | .98 | .68 | .83 |
| E. Macron | .91 | .94 | .52 | .71 | .95 | .96 | .59 | .79 |
| G. Meloni | .83 | .97 | .85 | .92 | .91 | .94 | .62 | .82 |
| N. Zingaretti | .90 | .93 | .62 | .78 | 1 | 1 | .72 | .90 |

The results in Tables 2 and 3 show on average a high agreement index for the variable *Presupposition Trigger* and, as expected, a much lower agreement for *Discourse Function* in both the politicians' and the ChatGPT texts. Due to the lower values of *Discourse Function*, the first and second author had a second *consensus through negotiation* meetings, in which each case of disagreement was discussed. Following this meeting, it was possible to produce a final dataset on which the quantitative analysis (Section 5.2) was performed.

## 5.2 Descriptive and inferential statistics

Regarding the frequency of PMPs in both politicians' speeches and ChatGPT texts (Q1, Section 3), Figure 1 shows that the absolute frequency of PMPs is on average higher in ChatGPT texts than in politicians' speeches.[9] The only exception seems to be M. Le Pen. However, the normalized frequencies shown in Figure 2 clearly show the higher frequency of PMPs in all AI-generated texts.

This first impression is confirmed by inferential statistics, which show that the distribution of PMPs in the politicians' speeches and ChatGPT texts is significantly different, although the effect size is small ($\chi^2 = 10.28$, df = 3, p < .05; Cramér's V = .16).

---

[9] The figures in this section were generated using the R package *ggplot2* (v. 3.4.2; Wickham 2016).

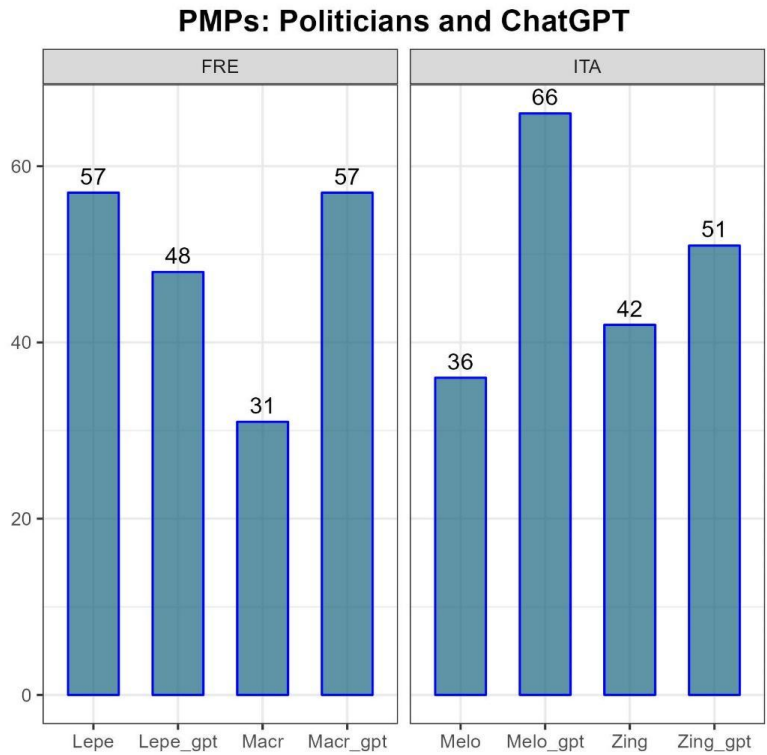

Figure 1: Absolute frequencies of PMPs in French and Italian politicians' text and ChatGPT's texts (n = 388).

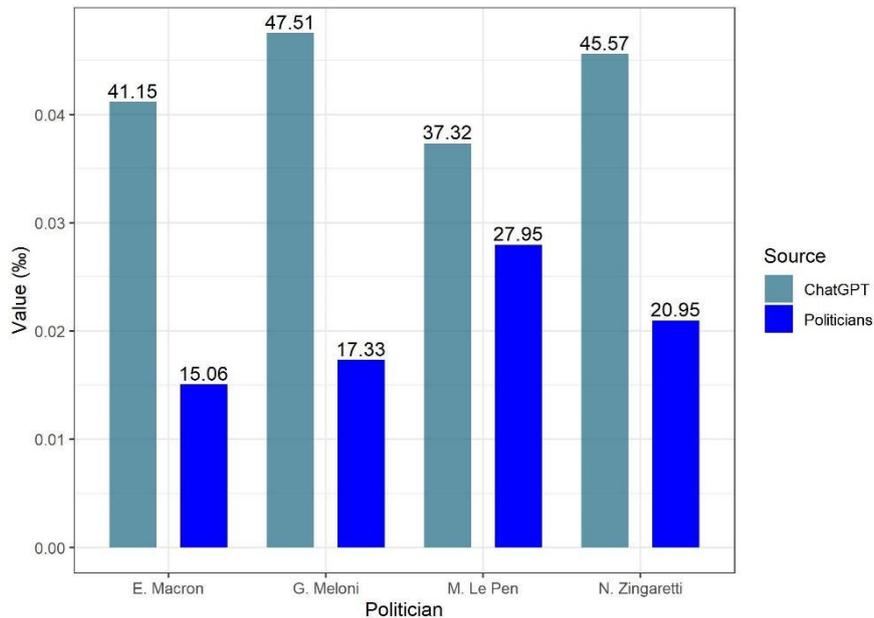

Figure 2: Normalized frequencies (per 1,000 words) of PMPs in French and Italian politicians' text and ChatGPT's texts.

Regarding the form of PMPs (Q2), we observe in Figure 3 that the most frequent presupposition triggers in both text types are change-of-state verbs (CSV) and definite descriptions (DEF), which together represent 70% of all triggers in the politicians' speeches and 80% of all triggers in the ChatGPT's texts. Other types of triggers are much less frequent (such as (semi-)factives, FCT, and relative clauses, REL) or almost unattested in the corpus (adjectives, ADJ, adverbs, ADV, cleft sentences, CLT, etc.).

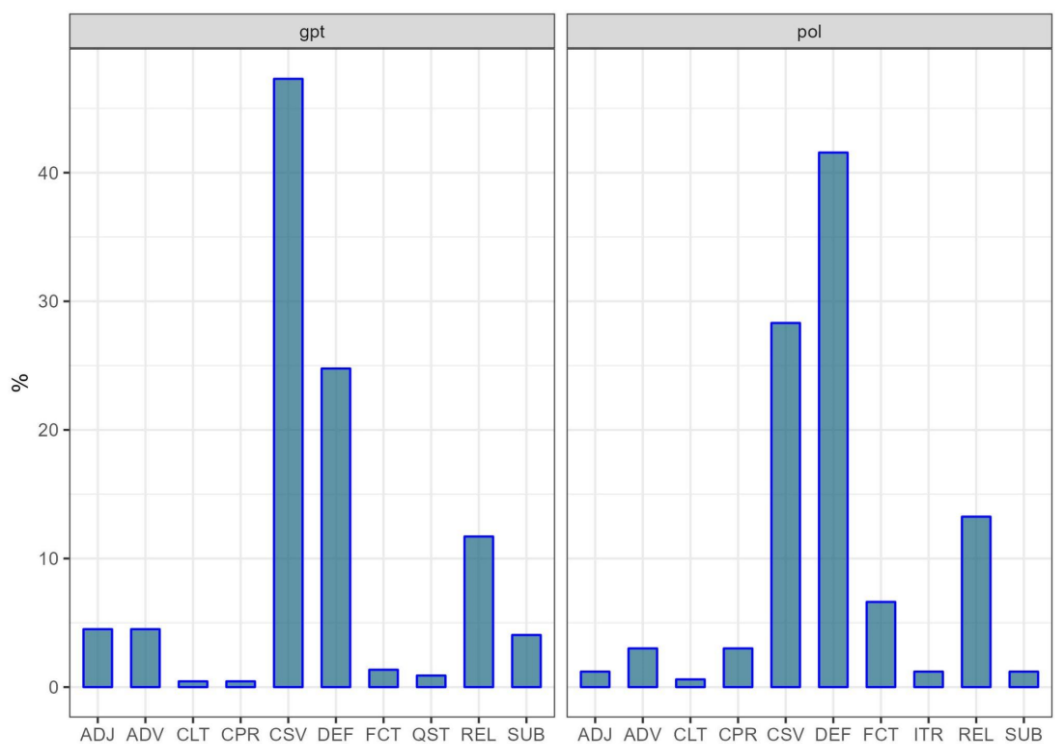

Figure 3: Presuppositions triggers (percentage) for PMPs in French and Italian politicians' and ChatGPT texts (ADJ = adjectives; ADV = adverbs; CLT = cleft sentences; CPR = comparatives; CSV = change-of-state verbs; DEF = definite descriptions; FCT = (semi-)factives; ITR = iterative verbs; QST = questions; REL = relative clauses; SUB = subordinate adverbial clauses).

Interestingly, the distribution of change-of-state verbs and definite descriptions, the most frequent presupposition triggers in the corpus, differs significantly in politicians' and ChatGPT texts with a moderate effect size ($\chi^2$ = 16.13, df = 1, p < .0001; Phi = .25). Definite descriptions are the most common way to convey a PMP in politicians' speeches, while change-of-state verbs are the most common in ChatGPT texts (see the discussion in Section 6).

Finally, regarding Q3, the distribution of the discourse functions differs significantly in politicians' and ChatGPT texts with a moderate effect size ($\chi^2$ = 49.19, df = 3, p < .0001; Cramér's V = .36). Stance-taking (STK) is the most common function for PMPs in both cases, but its overall proportion is much higher in the ChatGPT data. Furthermore, we observe a decrease in the frequency of PMPs with a Criticism function (CRT) when moving from the politicians' speeches to the ChatGPT texts.

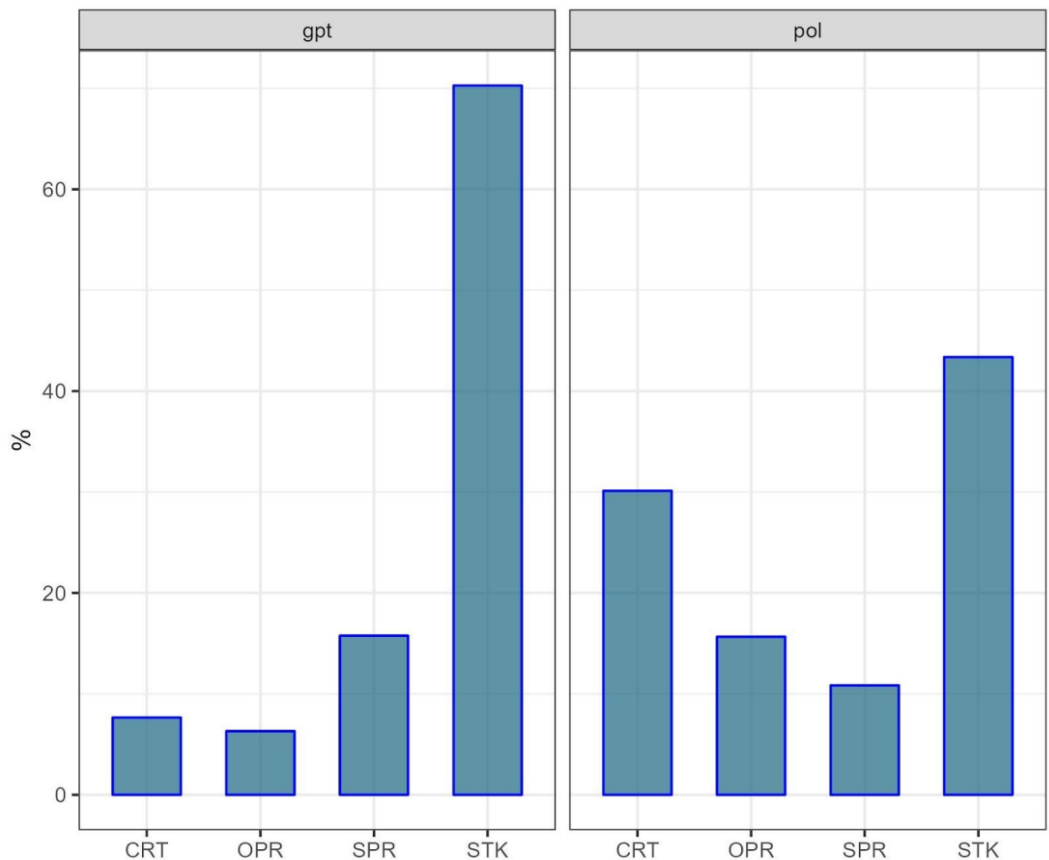

Figure 4: Discourse functions (percentage) conveyed by PMPs in French and Italian politicians' and ChatGPT texts (CRT = criticism; OPR = praise of others; SPR = self-praise; STK = stance-taking).

Focusing on individual variation (Figure 5), we observe that different politicians (left column) tend to associate with different discourse functions when they use PMPs. Even if Stance-taking is widespread overall, Praise of others seems to be more typical for M. Le Pen, Self-praise for E. Macron, and Criticism for G. Meloni. Looking at the ChatGPT data (right column), Stance-taking is massively represented and is clearly the most dominant category overall. However, there seems to be some variation. This seems particularly evident in Macron-GPT data, where Self-praise is almost as common as Stance-taking.

Finally, if we compare the distribution of the discourse functions in the politicians' and the chatbot's texts, we find significant differences in the case of M. Le Pen (Fisher's Exact Test, p < .001), G. Meloni (Fisher's Exact Test, p < .0001) and N. Zingaretti (Fisher's Exact Test, p < .01). The only notable exception is E. Macron's data: in this case, both his own speeches and those generated by ChatGPT show a similar distribution of discourse functions (Fisher's Exact Test = .93).

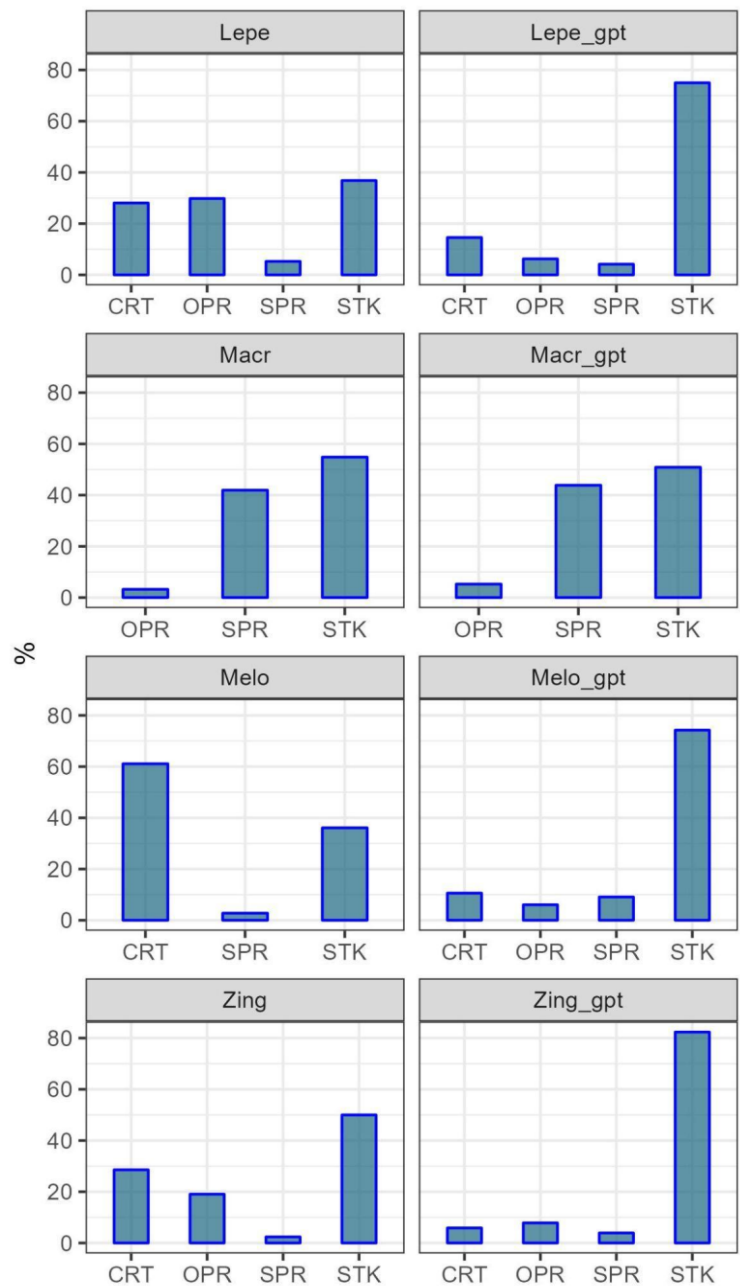

Figure 5: Percentage of *Discourse Functions* conveyed by PMPs in French and Italian politicians' text and ChatGPT's texts (CRT = criticism; OPR = praise of others; SPR = self-praise; STK = stance-taking). Individual variation.

To test the (in)dependence of more than two variables and their interactions simultaneously, we also performed a loglinear analysis, an extension of the Chi-square test for contingency tables with three or more categorical variables (for a detailed introduction, see Agresti 2019: 193-226; Field, Miles and Field 2012: 829–851).[10] For the present analysis, the variables *Discourse Function* (CTR; OPR; SPR; STK), *Group* (ChatGPT; Politicians) and *Language* (French; Italian) were

---

selected. Loglinear analysis does not require specifying a dependent variable, as the model predicts "the frequency of cases in different combinations of the predictors" (Field, Miles and Field 2012: 841). In other words, unlike other techniques such as logistic regression, loglinear analysis does not model an outcome (or dependent variable) as a function of one or more factors (the independent variables), but tests the association between different variables. Loglinear analysis involves stepwise backward elimination starting with a saturated model that includes all main effects and interactions, until the most parsimonious model that does not deviate significantly from the data is obtained (Field, Miles and Field 2012: 844). The formula of the final model is given in (5):

(5) ~ Group + Language + Discourse Function + Group: Discourse Function + Language: Discourse Function

The model in (5) retained all two-way interactions except the one between *Group* and *Language*. The likelihood ratio of this model was $\chi^2 = 8.14$, df = 4, p = .86. This model contains two significant interactions, which is the one between *Group* and *Discourse Function* ($\chi^2 = 50.99$, df = 3, p < .0001; Cramér's V= .29) and the one between *Language* and *Discourse Function* ($\chi^2 = 32.27$, df = 3, p < .0001; Cramér's V= .37).

To better interpret these two interactions, we can rely on the mosaic plots in Figures 6 and 7.[11] These plots provide a visual representation of the Standardized Residuals, which can be described (somewhat simplistically) as the difference between the observed and expected frequencies for each cell in the contingency table (see Agresti 2019: 39–41 for a detailed description). The color of the boxes indicates the sign of the residuals (positive, blue, when the observed frequency is higher than the expected frequency; negative, red, in the opposite case), while the color shades indicate the relative importance of a residual (as a rule of thumb, "the more intensive, the greater the deviation", Levshina 2015: 219). The area of the boxes reflects the proportion of the cells, rows and columns.

---

[11] These figures were generated using the R package *vcd* (v. 1.4-11; Zeileis, Meyer and Hornik 2007).

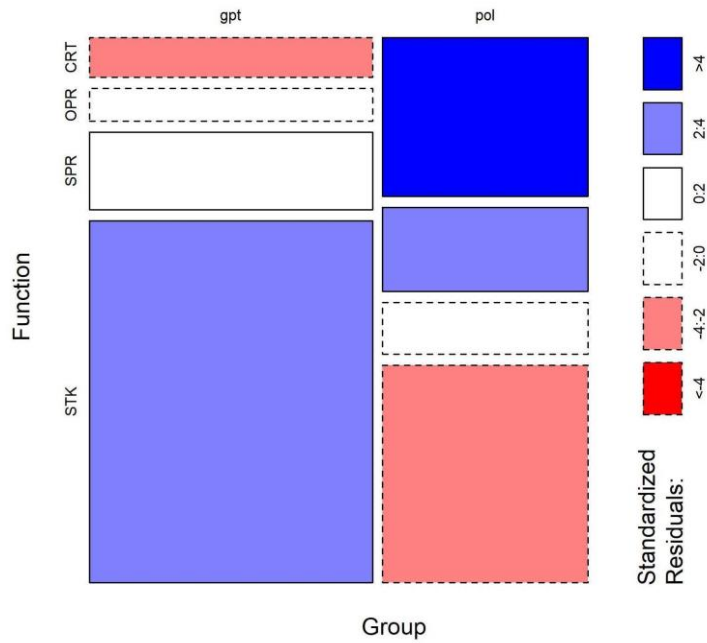

Figure 6: Interaction between the variables *Group* (ChatGPT, politician) and *Discourse Function* (CRT = criticism; OPR = praise of others; SPR = self-praise; STK = stance-taking).

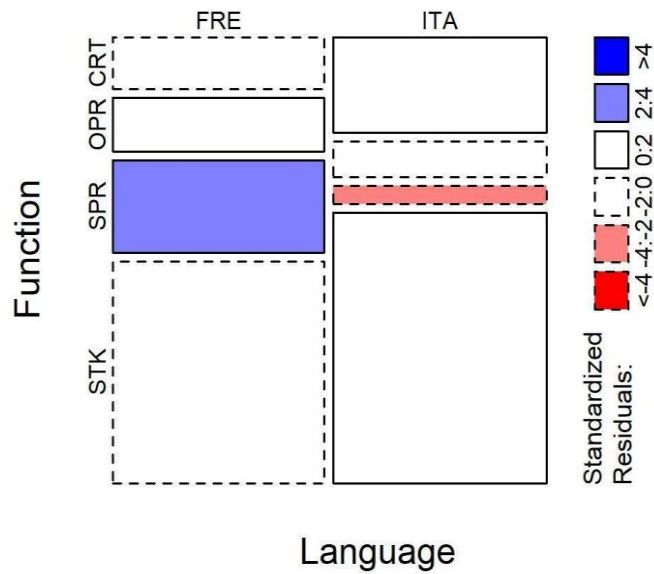

Figure 7: Interaction between the variables *Language* (French, Italian) and *Discourse Function* (CRT = criticism; OPR = praise of others; SPR = self-praise; STK = stance-taking).

These plots seem to confirm our previous observations, as we see in Figure 6 a significant positive association between Criticism and Praise of others with the politicians' texts (mainly thanks to G. Meloni and M. Le Pen, respectively; see Figure 5). At the same time, we also observe a significant positive association between Stance-taking and ChatGPT data as well as a significant negative association between Criticism and ChatGPT data.

In Figure 7, we can see that there are not many differences between the French and Italian data (regardless of whether they come from the politicians' texts or from ChatGPT), except when we consider Self-praise. In this case, we observe a significant positive association between Self-praise and French as well as a negative association between Self-praise and Italian. As we saw in Figure 5, this is mostly due to the overrepresentation of this category in the data of E. Macron, both in his speeches and in his chatbot counterpart.

## 6 Discussion

Coming back to our research questions, concerning Q1, we can observe that PMPs are usually more common in texts generated by LLMs.

Regarding Q2 and, in particular, the use and distribution of presupposition trigger types, change-of-state verbs are more frequent in the generated texts, whereas definite descriptions are more typical in the politicians' texts.

Finally, from a functional point of view, on average PMPs in ChatGPT texts are mainly used to convey a Stance-taking function, whereas in politicians' speeches they show a more diversified distribution and an overall positive association with Criticism, in contrast to ChatGPT. How can these differences be explained?

In our opinion, two patterns can be observed in the generated texts: a pattern of repetition and a pattern of vagueness. First, there is a tendency of ChatGPT to produce and repeat similar PMPs across languages and politicians, mainly by change-of-state state verbs, as in the examples (6) and (7):

(6) a. Nous devons travailler ensemble, main dans la main, pour *bâtir un avenir meilleur pour tous.* (Macron_ChatGPT, T2)
    'We must work together, hand in hand, to build a better future for all.'

   b. Siamo quelli che non si arrendono, che non si accontentano di parole vuote, ma che agiscono concretamente per *costruire un futuro migliore per tutti noi.* (Meloni_ChatGPT, T1)
   'We are the ones who do not give up, the ones who are not satisfied with empty words. We are the ones who take concrete action to build a better future for all of us.'

   c. Dobbiamo lavorare insieme *per costruire un futuro migliore per le prossime generazioni*, un futuro basato sulla solidarietà, sull'uguaglianza e sulla giustizia. (Zingaretti_ChatGPT, T1)
   'We must work together to build a better future for the next generation, a future based on solidarity, equality and justice.'

   d. Ensemble, *nous pouvons bâtir un avenir meilleur pour la France,* un avenir où notre nation retrouve sa fierté et son rayonnement. (Le Pen_ChatGPT, T1)
   'Together, we can build a better future for France, a future in which our nation regains its pride and influence.'

(7) a. Car je crois en notre peuple, en notre avenir, et en *notre capacité à reprendre notre destin en main* (Le Pen_ChatGPT, T2)
'Because I believe in our people, in our future, and in our ability to take our destiny back into our own hands.'

b. È tempo di *riprendere il controllo del nostro destino* (Meloni_ChatGPT, T2)
'It is time to regain control of our destiny.'

Example (6), in particular, is emblematic because the same presupposition trigger (the change-of-state verb *bâtir / costruire*, 'to build') appears in the texts of all four politicians in the corpus, regardless of their language and political orientation. Similar examples involve other change-of-state predicates such as *difendere, protéger,* 'to defend, to protect', *riconquistare,* 'to regain', *renforcer,* 'to strengthen', *retrouver,* 'to retrieve', etc.

Regarding the aforementioned pattern of vagueness, ChatGPT tends to produce less explicit criticism associated with PMPs, unlike politicians, see the examples in (8) (see also (3a)), all targeting the European Union:

(8) a. A *questa Europa senz'anima*, fatta solo a uso e consumo di alcuni (Meloni, T1)
'To this soulless Europe, only created for the use and consumption of a few.'

b. Et dans *ce mouvement de domination collective* [the European Union, ndA] sur les peuples, *les européistes sans repère* n'ont cessé d'avaliser un déséquilibre entre la France et l'Allemagne (Le Pen, T1)
'And in this movement of collective domination over the peoples, the Europeanists without reference points have constantly endorsed an imbalance between France and Germany.'

c. Il est temps [...] de mettre fin à *cette emprise de l'Union européenne qui menace notre identité nationale* (Le Pen_ChatGPT, T2)
'It's time [...] to put an end to the stranglehold of the European Union, which is threatening our national identity.'

In light of our analysis, we can claim that while politicians' speeches and their chatbot counterparts may be superficially similar, at a deeper level they differ significantly in terms of the distribution of PMPs and their discourse functions. The differences shown by ChatGPT texts may be related to known biases and limitations of LLMs. In particular, ChatGPT texts are more general and vague (cf. Antoun, Mouilleron, Sago and Seddah 2023) and show a stronger tendency towards verbosity and repetition of certain phrases (Pratim Ray 2023; De Cesare 2023) than the politicians' speeches. This may explain why PMPs are more frequent in ChatGPT texts and why they are mostly conveyed by change-of-state verbs (see Figure 3, Section 5.2.): as we have seen in (6) and (7), ChatGPT tends to use and repeat the same noun phrases formed by change-of-state predicates as verbal heads and direct objects (e.g., *costruire il futuro*, 'building our future') within and across texts, resulting in a higher percentage of PMPs. The reason for this is likely linked to the architecture of ChatGPT. This chatbot generates sentences "by predicting upcoming words, evaluating the probabilities of potential candidates as a function of previously produced words but not yet-to-be-produced words" (Cai, Duang and Haslett 2023: 18). It is therefore possible that the LLM, on the basis of its training data, recognizes sentences such as those examined in (6) and (7) as particularly likely and salient in this text genre.

The exception posed by E. Macron's data could depend on the training data (which are inaccessible) or on the excerpt used in the part c) of the prompt. This prompt contains indeed different examples of the Self-praise function (extracted from Macron, T2), as is evident in (9):

(9) Grâce aux réformes menées, notre industrie a pour la première fois recréé des emplois et le chômage a atteint son plus bas niveau depuis quinze ans. Grâce au travail de tous, nous avons pu investir dans nos hôpitaux et notre recherche, *renforcer nos armées*, recruter policiers, gendarmes, magistrats et enseignants, *réduire notre dépendance aux énergies fossiles*, *continuer à moderniser notre agriculture* (Macron, T2).
'As a result of our reforms, our industry has created new jobs for the first time and unemployment has fallen to its lowest level in fifteen years. Thanks to everyone's hard work, we have been able to invest in our hospitals and research, strengthen our armed forces, recruit police, gendarmes, judges and teachers, reduce our dependence on fossil fuels and continue to modernize our agriculture.'

However, a potential bias due to the high frequency of a particular function in the prompt is not replicated in the case of Meloni_ChatGPT, T1. The prompt contains an excerpt that includes several PMPs with the discourse function Criticism, but the text generated by the LLM does not show an overrepresentation of this type of PMP (see OSF).

Interestingly, though, as also suggested by other work (Cai, Duang and Haslett 2023; Barattieri di San Pietro, Frau, Mangiaterra and Bambini 2023; Qiu, Duan and Cai 2023), ChatGPT shows pragmatic competencies analogous to human agents (at least under certain conditions). This seems to be the case for the ability to make appropriate use of PMPs, albeit with a higher frequency compared to politicians. To some extent, this also suggests that LLMs are sensitive to the effectiveness of certain strategies that rely on implicit meaning in political communication.

## 7 Conclusion

In this paper, we examined the use of LLM-generated texts in replicating pragmatic devices, in particular *non-bona fide* true presuppositions (PMPs), commonly used in persuasive speeches such as those used in political rallies. Our analysis revealed the ability of LLMs such as ChatGPT to mimic implicit linguistic nuances and create identities that match the subtleties of real politicians' argumentative styles.

A detailed investigation highlighted the potential for AI to generate potentially manipulative presuppositions through vague, generic slogans. This research intended to provide insights into the expressive creativity of LLMs and their potential role in the creation of persuasive narratives. More generally, the paper contributes to the knowledge of how generative AI can also emulate human language in pragmatic aspects (Barattieri di San Pietro, Frau, Mangiaterra and Bambini 2023; Qiu, Duan and Cai 2023).

It is also crucial to address two primary limitations of our study. First, the corpus used for the analysis may not be representative; the size of our manually annotated data was restricted, posing challenges in fully realizing a corpus-based approach. This limitation reflects an inherent trade-off when dealing with pragmatic categories that currently elude automatic detection. This limited size could lead to biased results, especially for the politicians' corpus. While the linguistic patterns of ChatGPT are based on a very large set of examples drawn from political communication, thus making it *de facto* representative, the use of PMPs in our small politicians' corpus is inevitably bound to the contingent text selection. Second, the stochastic nature of LLMs in text generation introduces variability into the results, implying that the generated texts may not be all the way replicable. However, manually setting parameters such as *temperature* and *diversity_penalty* can help standardize the output. According to ChatGPT, users cannot adjust these parameters on a per-request basis in the ChatGPT chat interface since they are pre-configured by the model.[12] However, our trials show that the responses varied according to changing parameters in the web input, as other users also observed.[13] The results obtained from the input in the chat interface and the same prompt accessing ChatGPT API from R were almost identical as far as presuppositions are concerned. This is likely due to the fact that our manually adjusted parameters coincide with the parameters configured by the model. However, the use of APIs for parameter insertion could improve the control of prompts and eventually generate more specific responses. Finally, we should not forget that the texts generated by ChatGPT for this study are in French and Italian, while the vast majority of the training data for this LLM is in English (93% of the data for GPT 3 was set in English, see Brown, Mann, Ryder, Subbiah et al. 2020: 14). This large imbalance in the ChatGPT training dataset may therefore be an unpredictable source of English interference when generating texts in other languages.

Despite all these limitations, our study provides a contribution to exploring the pragmatic aspects of AI-generated texts, particularly in identifying non-*bona fide* true presuppositions and their alignment with real-world political communication. As a next step, it would be intriguing to explore how advancing AI technologies, such as ChatGPT 4.0, might further refine the detection and application of these nuances, also extending our analysis to other strategies of implicit communication (such as vagueness). In addition, a more detailed analysis of the use of these PMPs, considering not only the choice of change-of-state verbs but also their specific referential targets would enrich our understanding of the distribution of PMPs in ChatGPT-generated texts. In conclusion, our research underscores the growing sophistication of AI in mimicking complex human communication patterns, which presents both opportunities and challenges for the future of AI-driven persuasive language.

---

[12] https://community.openai.com/t/does-the-chatgpt-api-parameters-also-work-in-the-chatgpt-web-version/368454

[13] https://www.linkedin.com/pulse/temperature-check-guide-best-chatgpt-feature-youre-using-berkowitz

## Attributions

**Davide Garassino**: Conceptualization, methodology, investigation, resources, formal analysis, data curation, writing-original, visualization, review and editing. In particular, he wrote: §1, §2.1, §3, §5.1, §5.2, and, jointly with Viviana Masia, §6. **Viviana Masia**: Conceptualization, data curation, investigation, writing-original, review and editing. In particular, she wrote: §2.2, and, jointly with Davide Garassino, §6. **Nicola Brocca**: Conceptualization, software, investigation, data-curation, writing-original, review and editing. In particular, he wrote: §4 and §7. **Alice Delorme Benites:** Conceptualization.